\documentclass[letterpaper, 10pt, conference]{ieeeconf} 
\IEEEoverridecommandlockouts 
\usepackage{graphicx}

\usepackage{indentfirst}
\usepackage{array,multirow}
\usepackage{fancyhdr}
\usepackage{float}
\usepackage{booktabs}
\usepackage{multirow}
\usepackage{here}
\usepackage{svg} 
\usepackage{comment}
\usepackage{subfigure}
\usepackage{caption}
\usepackage{siunitx}
\usepackage{hyperref}
\usepackage{algorithmicx,algorithm,algpseudocode}
\usepackage{amsmath,amssymb,amsfonts,bm,textcomp}
\usepackage{fontawesome5}
\usepackage{cite}

\title{\LARGE \bf
Cooperative Path-following Control of Remotely Operated 
\\
Underwater Robots for Human Visual Inspection Task
\vspace{-2mm}}

\author{
Eito~Sato$^{1}$, Hailong~Liu$^{1}$, Norimitsu~Sakagami$^{2}$ and Takahiro~Wada$^{1}$
\thanks{$^{1}$Eito~Sato, Hailong~Liu, and Takahiro~Wada with the Graduate School of Science and Technology, Nara Institute of Science and Technology, 8916-5 Takayama-cho, Ikoma, Nara 630-0192, JAPAN. 
{\tt\small \{sato.eito.sd7; liu.hailong; t.wada\}@is.naist.jp}}
\thanks{$^{2}$Norimitsu~Sakagami with the department of Navigation and Ocean Engineering, Tokai University, Shizuoka, JAPAN.
{\tt\small sakagami@scc.u-tokai.ac.jp}}%
}

\begin{document}
\maketitle
\thispagestyle{plain}
\pagestyle{plain}

\begin{abstract}
Remotely operated vehicles (ROVs) have drawn much attention to underwater tasks, such as the inspection and maintenance of infrastructure. The workload of ROV operators tends to be high, even for the skilled ones. Therefore, assistance methods for the operators are desired. This study focuses on a task in which a human operator controls an underwater robot to follow a certain path while visually inspecting objects in the vicinity of the path. In such a task, it is desirable to realize the speed of trajectory control manually because the visual inspection is performed by a human operator. However, to allocate resources to visual inspection, it is desirable to minimize the workload on the path-following by assisting with the automatic control. Therefore, the objective of this study was to develop a cooperative path-following control method that achieves the above-mentioned task by expanding a robust path-following control law of nonholonomic wheeled vehicles. To simplify this problem, we considered a path-following and visual objects recognition task in a two-dimensional plane. We conducted an experiment with participants (n=16) who completed the task using the proposed method and manual control. The results showed that both the path-following errors and the workload of the participants were significantly smaller with the proposed method than with manual control. In addition, subjective responses demonstrated that operator attention tended to be allocated to objects recognition rather than robot operation tasks with the proposed method. These results indicate the effectiveness of the proposed cooperative path-following control method.
\end{abstract}

\section{ INTRODUCTION}

Unmanned underwater vehicles (UUVs) have drawn much attention owing to the rapid increase in the demand for underwater tasks~\cite{01_shimono2016development}.
UUVs are classified into autonomous underwater vehicles~(AUVs) and remotely operated vehicles~(ROVs).
AUVs are controlled by using all the sensor information acquired by the underwater robot. 
AUVs can be operated for a long time, and are often used for topographic surveys and water quality testing~\cite{03_hwang2019auv}.
However, AUVs have some limitations~\cite{26_daoliang2021auv}, such as the need for a specific environment to be prepared in advance~\cite{f} and the difficulty in performing complex movements~\cite{g}. 
Therefore, it is difficult to rely solely on autonomous control by the system as it is currently limited to specific cases and is difficult to deal with unexpected situations, such as disturbances.
Human operators are required to frequently participate in some underwater tasks~\cite{b,c}, such as the inspection and maintenance of underwater infrastructure. 
Thus, ROVs which are controlled by a human operator are widely used for those tasks.

In this study, we focused on underwater inspection along a certain trajectory or path using ROVs.
It has been pointed out that operation training is indispensable for ROV operators to achieve satisfactory performance because advanced skill is required. 
For example, the operator needs to operate the ROV based on a limited source of information, such as video images from a camera attached to the ROVs and/or poor perception of the position and orientation of the robot. 
The workload of the operators also tends to be high even for the skilled operators
~\cite{02_azis2012problem, 24_ho2011humanfactor,27_manual2014}.
Therefore, assistance methods for the ROV operators have been desired for improving the work efficiency, decreasing the workload, and prolonging the work time of the operator.

Cooperative control, which is a combination of automatic control and human manual control, is a promising approach to assist human operators in a variety of remotely operated or other types of human–machine systems~\cite{abbink2,22_wada2019simulation} because it can incorporate advanced technologies of autonomous control and/or other machine intelligence with manual control. 
Shared control is one of the cooperative control schemes, in which humans and machines interact congruently in a perception-action cycle to jointly perform a dynamic task\cite{06_mulder2012sharing}. 
Haptic shared control~(HSC) is a branch of the shared control in which control inputs of both humans and a machine are force or torque to a single control terminal, and the human operator perceives the input of the machine through the haptic information sensation generated by the terminal\cite{04_abbink2012haptic,05_nishimura2015haptic}. 
HSCs have been introduced in a wide range of fields of human–machine cooperation, such as the automotive field to assist drivers~\cite{25_amir2019drive}, telemanipulation in space~\cite{07_kimmer2015effects}, underwater robots \cite{08_konishi2020haptic,09_kuiper2013haptic}, and aircraft control~\cite{10_lam2007haptic}. 
All of the above studies suggest that the total system performance improves on account of human–machine interaction.
Cooperative control methods are designed according to the content of a given task, in which different cooperative schemes are combined in some cases.

The present study focuses on a task in which a human operator remotely controls an underwater robot to follow a certain path while visually inspecting objects in the vicinity of the path. 
Such tasks are found in underwater missions, such as pipe inspection\cite{23_kim2020pipe}.
In such a task, it is desirable to realize the speed of trajectory control manually because visual inspection is performed by a human. 
However, to allocate resources to visual inspection, it is desirable to minimize the workload on the task of path-following by assisting with automatic control. 
Therefore, the purpose of this study is to develop a cooperative path-following control method for ROVs suitable for human visual inspection to improve the path-following performance of the overall system and reduce the workload of the operator who conducts visual inspection and path-following by remotely operating the robot simultaneously.
Specifically, we extend the path-following control of nonholonomic wheeled vehicles based on the inverse optimal control law~\cite{12_Kurashiki2007} to holonomic underwater robots and propose a new cooperative path-following control method that is suitable for human visual inspection tasks by combining the path-following controller and operator guidance methods. 

In the proposed method, HSC is applied to guide the direction of the heading angle control of the robot. 
For translational motion, the direction of the vehicle movement is automatically tuned according to the magnitude of the lateral deviation from the target path to further assist the path-following activity of the operator, while the speed of the translational motion is determined by the input of the operator.
The proposed method was implemented on an underwater robot, and experiments, in which participants performed an objects recognition task during path-following, were conducted to investigate the effectiveness of the proposed method.

This paper is organized as follows.
Section~\ref{proposed} describes the design concept as well as the detail of the proposed method.
Section~\ref{experiment} describes the experimental method.
In Section~\ref{result}, the results of the subjective experiment are presented.
Section~\ref{discussion} describes the discussion of the obtained results. 
Finally, conclusions are given in Section~\ref{conclusion}.

\section{PROPOSED METHOD}
\label{proposed}

\subsection{Design of Cooperative Control}

In the present study, we propose a new cooperative path-following control scheme that includes two guidance methods to reduce the operator workload and improve the path-following accuracy.

\begin{figure}[tb]
\vspace{2mm}
\centering
\includegraphics[width=0.9\linewidth]{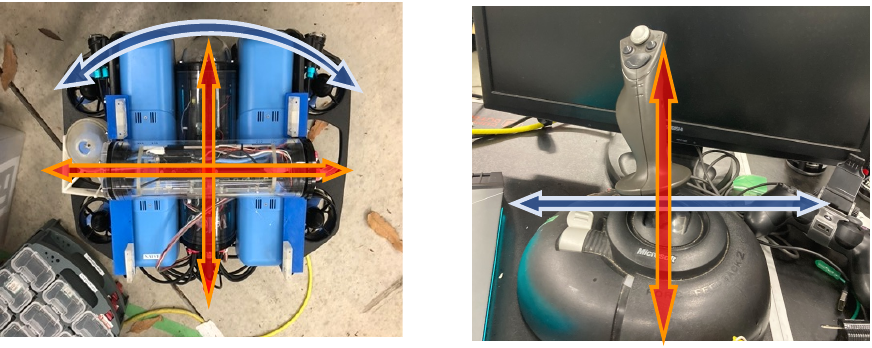}
\caption{Correspondence between the joystick and robot.}
 \label{fig:dimesion}
  \end{figure}

When the underwater robot moves in a two-dimensional plane, the motion of the robot with three degrees of freedom (DoF), as shown in the left part of Fig.\ref{fig:dimesion} needs to be specified.
However, the operation of all three-DoF movements is expected to be difficult for novice operators to master. 
Thus, it is considered that the three-DoF movement is reduced to two-DoF movement so that the operator can operate more intuitively with a lower workload.
Among several ways to degenerate dimensions, the most commonly used method is to assign the longitudinal and lateral directions of the joystick to the translational and rotating motions of the underwater robot, respectively~(Fig.~\ref{fig:dimesion}).
When an underwater robot is used for objects recognition or visual inspection while following a path, it is important to not change the robot orientation significantly. 
Thus, the dimension for the orientation controls is assigned to the independent joystick motion.

The proposed assistance method provides two different guidance methods developed for translational and rotational motions of the underwater robots, which are described in the following section:
\begin{enumerate}

\item [1)] Translational velocity conversion: It is the first guidance method employed to automatically determine the direction of translational motion in which the underwater robot travels. 
The operator determines only the magnitude of the translational velocity of the underwater robot by operating in the longitudinal direction of the joystick, and the system determines its direction according to the distance to the path. 
This guidance method cannot be interfered with by the operator only by the longitudinal joystick motion, while the operator can resist this guidance by changing the orientation of the robot using lateral joystick motion.

\item [2)] Haptic guidance: The second guidance is that the HSC indicates how the operator should operate the joystick to correct the posture of the underwater robot.
This guidance provides the operator with haptic information on how to operate the joystick in the lateral direction to achieve the target angular velocity of the underwater robot using the control law described in section~\ref{control}.
This guidance is mainly beneficial for following a curved path or maintaining the posture to align the direction of the path to stabilize the video image observed by the operator for visual inspection.
The operator can ignore or defy the haptic guidance to operate the underwater robot as the operator desires.
The operator is allowed to defy the guidance because it can overcome the situation in which the recognition of the desired path is incomplete by responding flexibly. 

\end{enumerate}

\subsection{Cooperative Path-following Control for Holonomic Underwater Robots}
\label{control}

 \begin{figure}[tb]
\vspace{2mm}
\centering
\includegraphics[width=0.8\linewidth]{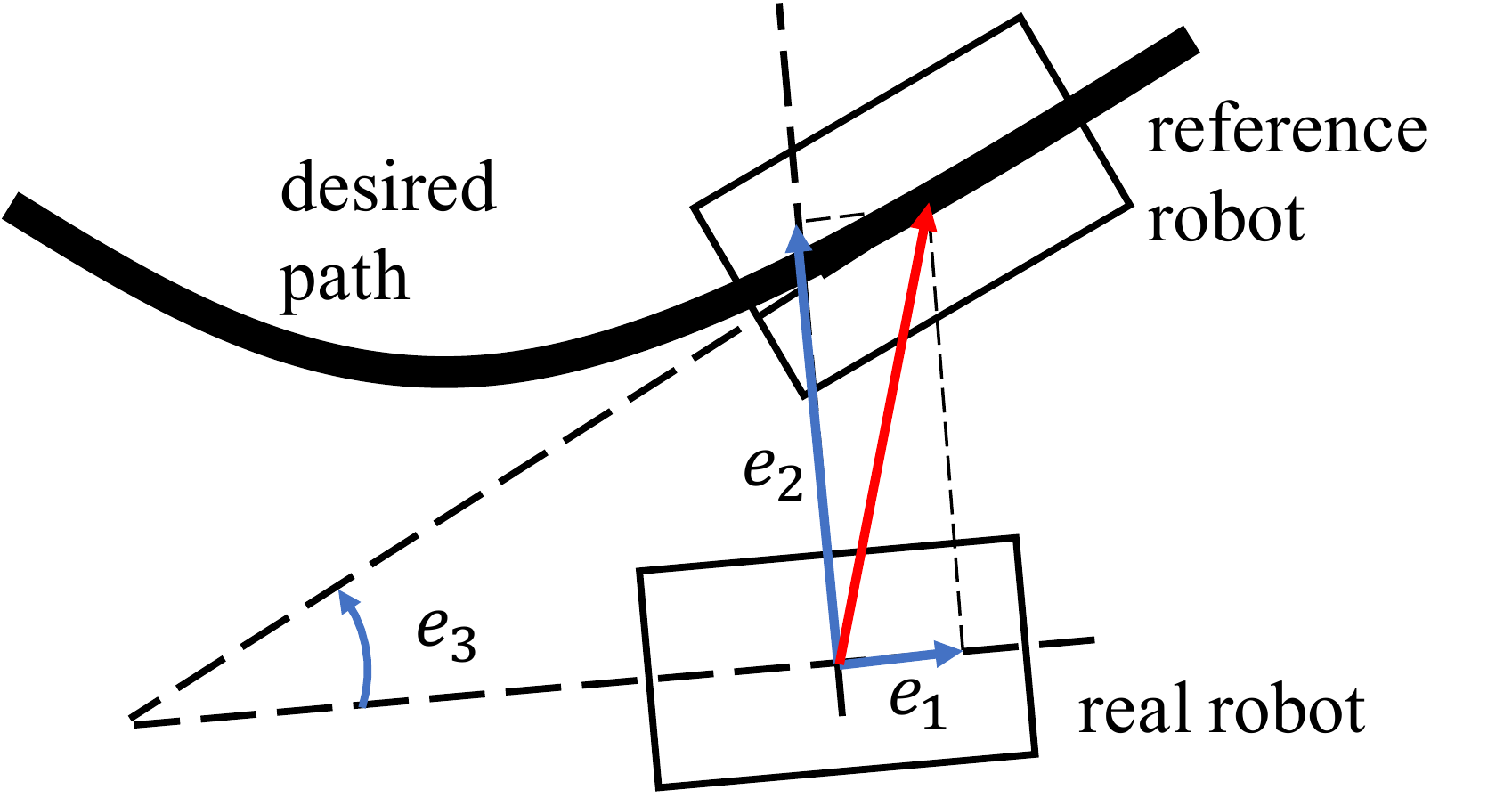}
\caption{Definition of the Path-following error model~\cite{12_Kurashiki2007}.}
 \label{fig:model}
\centering
\includegraphics[width=0.70\linewidth]{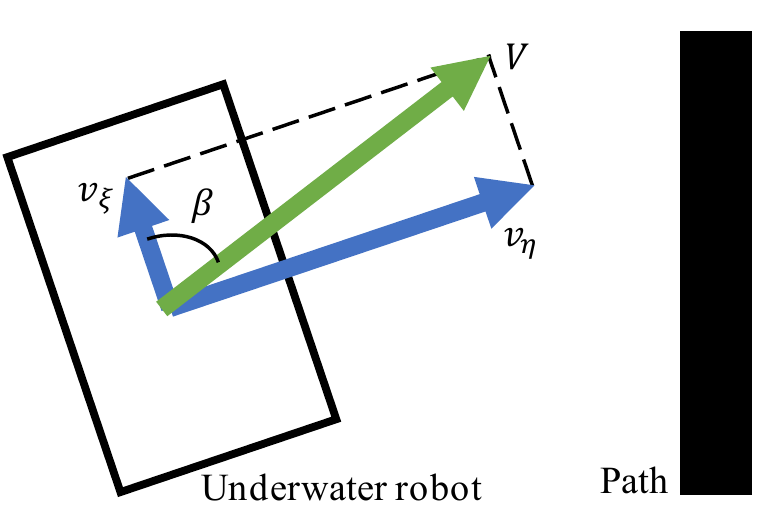}
\caption{Determination method of movement direction ($\beta$).}
 \label{fig:tranclational velocity}
\vspace{-2mm}
\end{figure}

\begin{figure*}[ht]
\vspace{2mm}
\centering
\includegraphics[width=0.95\linewidth]{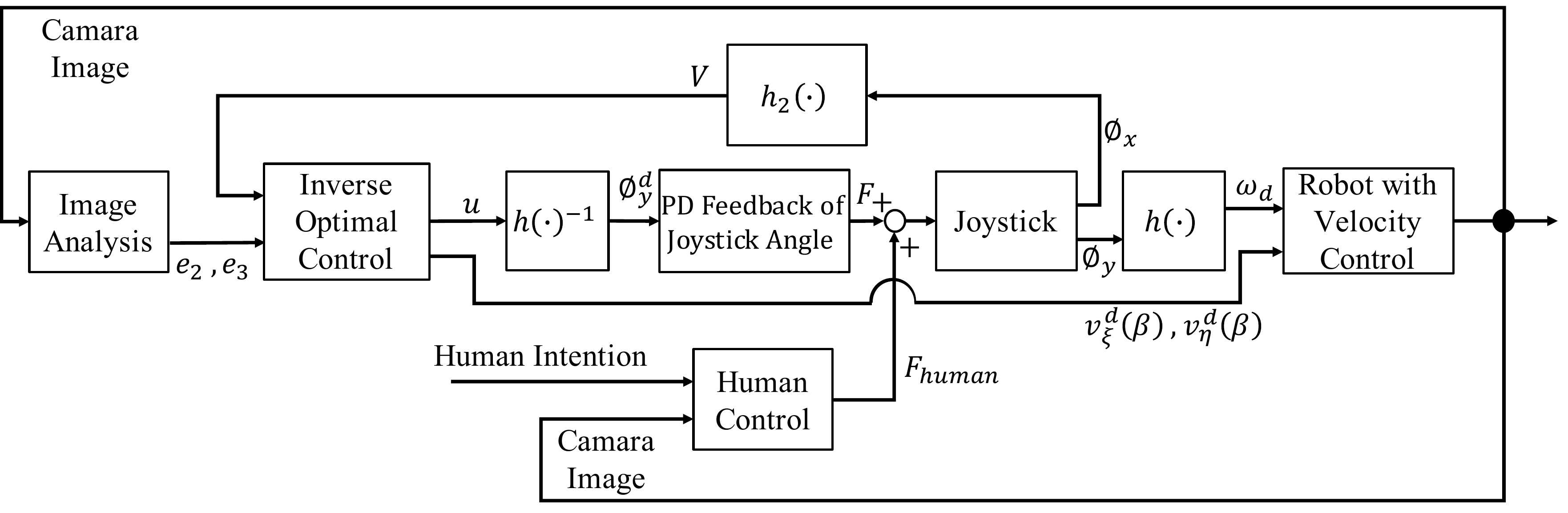}
\caption{Block diagram of the proposed method used for an underwater robot.}
 \label{fig:block diagram}
 \vspace{-2mm}
\end{figure*}

A new cooperative path-following control method for holonomic vehicles suitable for cooperation with humans is proposed by extending the path-following controller of nonholonomic wheeled vehicles using the inverse optimal control law~\cite{11_Araki2016,12_Kurashiki2007}. 
The first guidance method, which is the method of translational velocity conversion, is incorporated into the control law.

As shown in Fig.~\ref{fig:model}, we consider a reference robot moving on a desired path and design a control input that reduces the error between the real and reference robots. 
Although there is a delay between the input to the thrusters and the velocity of the underwater robot owing to the complexity of the system dynamics, we derive the dynamics of the controlled plant, in which the translational and angular velocities of the robot in the robot-fixed coordinate system are considered as the input by assuming the robot motion in the low-velocity range.

The position and orientation of the real robot in the world coordinate system $q={\left[x,y,\theta \right]}^T$, and the velocity in the robot-fixed coordinate system [$v_{\xi}, v_{\eta}, \omega$] satisfy~\eqref{eq:1}.
\vspace{-0mm}
\begin{eqnarray}
\frac{d}{dt}\left[ \begin{array}{c}
x \\ 
y \\ 
\theta  \end{array}
\right]&=&\left[ \begin{array}{ccc}
{\cos \theta \ } & -{\sin \theta \ } & 0 \\ 
{\sin \theta \ } & {\cos \theta \ } & 0 \\ 
0 & 0 & 1 \end{array}
\right]\left[ \begin{array}{c}
v_{\xi } \\ 
v_{\eta } \\ 
\omega  \end{array}
\right] ,
\label{eq:1}
\end{eqnarray}

\noindent where $v_{\xi}$, $v_{\eta}$, and $\omega$ denote the longitudinal, lateral, and angular velocities of the robot, respectively.

The position error between the real robot (i.e., $[x,y,\theta]$) and the reference robot (i.e., $[x_r,y_r,\theta_r]$), which has the same dynamics as the real robot, as shown in Fig.~\ref{fig:model}, satisfies the following equation:
\begin{eqnarray}
\left[ \begin{array}{c}
e_1 \\ 
e_2 \\ 
e_3 \end{array}
\right]&=&\left[ \begin{array}{ccc}
{\cos \theta \ } & {\sin \theta \ } & 0 \\ 
{\mathrm{-}\sin \theta \ } & {\cos \theta \ } & 0 \\ 
0 & 0 & 1 \end{array}
\right]\left[ \begin{array}{c}
x_r-x \\ 
y_r-y \\ 
{\theta }_r-\theta  \end{array}
\right]
\label{eq:3}
\end{eqnarray}

Differentiating~\eqref{eq:3} with time yields the following equation:
\begin{eqnarray}
\setlength{\arraycolsep}{0.75mm}{
\frac{d}{dt}\left[ \begin{array}{c}
e_1 \\ 
e_2 \\ 
e_3 \end{array}
\right]=\left[ \begin{array}{c}
v_{\xi r}{\cos e_3\ } \\ 
v_{\xi r}{\sin e_3\ } \\ 
{\omega }_r \end{array}
\right] + 
\left[ \begin{array}{ccc}
-1 & 0 & e_2 \\ 
0 & -1 & -e_1 \\ 
0 & 0 & -1 \end{array}
\right]\left[ \begin{array}{c}
v_{\xi } \\ 
v_{\eta } \\ 
\omega  \end{array}
\right],
}
\label{eq:5}
\end{eqnarray}
where $v_{\xi r}, v_{\eta r}$, and $\omega_r$ denote the longitudinal, lateral, and angular velocities of the reference robot, respectively, and $v_{\eta r}=0 $ is applied.

In this system, three variables, $v_\xi$, $v_\eta$, and $\omega$, can be regarded as the inputs at this stage. 
To realize cooperative control with humans, we consider degenerating the input from three dimensions into two dimensions. 
In the context of this study, it is assumed that it is important to allow the operators to decide the speed of the path-following and realize the process of path-following in a manner that does not change the orientation of the robot to avoid interfering with human visual inspection as much as possible.
To achieve this, we propose that the magnitude of the velocity $V$ is determined by the longitudinal lever input of the operator and that the direction of $V$ is determined by~\eqref{eq:7} and~\eqref{eq:6}. $V_r$ is the longitudinal velocity of the reference robot~(Fig.~\ref{fig:model}).

\begin{eqnarray}
&v_{\xi }=V{\cos \beta},\ v_{\eta}=V{\sin \beta },\ v_{\xi r}=V_r&\label{eq:7} \\
&\beta=-{\mathrm{\arctan} \left(\alpha e_2\right)}&
\label{eq:6}
\end{eqnarray}

The direction of translation is determined according to the lateral error $e_2$, as shown in Fig.\ref{fig:tranclational velocity}. Substituting~\eqref{eq:7} into~\eqref{eq:5} yields~\eqref{eq:8}:
\begin{eqnarray}
\setlength{\arraycolsep}{0.85mm}{
\frac{d}{dt}\left[ \begin{array}{c}
e_1 \\ 
e_2 \\ 
e_3 \end{array}
\right]=\left[ \begin{array}{c}
V_r{\cos e_3\ } \\ 
V_r{\sin e_3\ } \\ 
{\omega }_r \end{array}
\right]+\left[ \begin{array}{cc}
{\mathrm{-}\cos \beta \ } & e_2 \\ 
{\mathrm{-}\sin \beta \ } & -e_1 \\ 
0 & -1 \end{array}
\right]\left[ \begin{array}{c}
V \\ 
\omega  \end{array}
\right] .
}
\label{eq:8}
\end{eqnarray}
In the case of path-following, because longitudinal errors are not considered, $V_r$ is determined such that $e_1=0$ and $de_1/dt=0$, yielding the following equation:

\begin{eqnarray}
V_r=\frac{V{\cos \beta \ }-e_2\omega }{{\cos e_3\ }}
\label{eq:9}
\end{eqnarray}
The angular velocity of the reference robot ${\omega}_r$ is determined by $\omega_r=\rho V_r$,
where $\rho$ denotes the curvature of the path at the point where the reference robot is located.

According to the above discussion, the error dynamics of the controlled plant can be expressed by an input-affine system, as referenced in~\eqref{eq:8}, with state variables $\textbf{e}:={\left[e_2,e_3 \right]}^T$ and input $u:=\omega$
\begin{eqnarray}
\dot{\textbf{e}}\ =\ f\left(\textbf{e}\right)+\ g\left(\textbf{e}\right)u\label{eq:11}
\end{eqnarray}

\noindent where
\begin{eqnarray}
f\left(\textbf{e}\right) :=
\left[ \begin{array}{c}
V_r \sin{e_3}-V\sin{\beta} \\ 
\omega_r\end{array}
 \right] , 
g\left(\textbf{e}\right) :=
\left[\begin{array}{c}
0\\-1\end{array}\right].
\end{eqnarray}

The inverse optimal control \cite{12_Kurashiki2007} is designed for a given~\eqref{eq:11}. 
We consider the following positive definite function:
\begin{eqnarray}
V_0\left(\textbf{e}\right)=\frac{1}{2}K_2e^2_2+\frac{1}{2}K_3e^2_3.
\label{eq:12}
\end{eqnarray}
The $V_0\left(\textbf{e}\right)$ is proven to be the control Lyapunov function in the input-affine system of~\eqref{eq:11}, since it satisfies $L_g V_0\left(\textbf{e}\right)=0\Longrightarrow L_fV_0(\textbf{e})<0$ when $V\neq0$~\cite{13_fukao2004inverse}, in which 
$L_g V_0\left(\textbf{e}\right)$ and $L_fV_0(\textbf{e})$ donate Lie derivative $g(\textbf{e}) \frac{\partial V_0\left(\textbf{e}\right)}{\partial \textbf{e}}$, $f(\textbf{e}) \frac{\partial V_0\left(\textbf{e}\right)}{\partial \textbf{e}}$. 
Therefore, it can be shown that the control input in~\eqref{eq:13} asymptotically stabilizes the error $\textbf{e}=[e_2,e_3]^T$ to zero~\cite{12_Kurashiki2007,13_fukao2004inverse}.
\begin{eqnarray}
u=-p\left(\textbf{e}\right)b\left(\textbf{e}\right)\label{eq:13}
\end{eqnarray}
\begin{eqnarray}
a\left(\textbf{e}\right):=L_fV_0\left(\textbf{e}\right),\ b\left(\textbf{e}\right):=L_gV_0\left(\textbf{e}\right)
\label{eq:14}
\end{eqnarray}
\begin{eqnarray}
p\left(\textbf{e}\right):=\left\{ \begin{array}{l}
c_0+\frac{a\left(\textbf{e}\right)+\sqrt{a\left(\textbf{e}\right)^2+{\left(b\left(\textbf{e}\right)^Tb\left(\textbf{e}\right)\right)}^2}}{b\left(\textbf{e}\right)^Tb\left(\textbf{e}\right)} \left(b\left(\textbf{e}\right)\neq 0\right)\\
c_0 \left(b\left(\textbf{e}\right)=0\right)
\end{array}
\right.
\label{eq:15}
\end{eqnarray}

It should be noted that the controller can be temporarily turned off by setting the errors ($e_2, e_3$) to zero when the system determines that it does not recognize the path to be followed for any reason.
In such a case, $\beta$ in~\eqref{eq:6} and $u$ in~\eqref{eq:13} will be zero, and the assistance by the system does not work.
In addition, the operator also allows the assistance to be turned off by pushing a button when the operator recognizes that the system is silent in failure.

\subsection{Haptic Shared Control for Operating Guidance}

The second guidance method, operating guidance, informs the operator through the haptic sensation how the underwater robot should be turned by the HSC.
Fig.\ref{fig:block diagram} includes the detailed structure of the designed HSC.
The desired angular velocity $u$ calculated from the control law described in~\eqref{eq:13} is converted to the target angle $\phi_d$ of the joystick by using $h(\cdot)^{-1}$, where $h(\cdot)$ denotes the relationship between the desired angular velocity and joystick angle as follows:
\begin{equation}
\omega_d = h( \phi_y).
\label{eq:hsc}
\end{equation}
The haptic controller determines the force generated by the joystick in the lateral direction as $F=K_p(\phi_y - \phi_y^d)+K_d \dot{\phi_y}$.
In addition to this force $F$ by the haptic shared controller, the operator can also apply the force $F_{human}$ to the joystick, and the combined force determines the joystick motion $\phi_y$. 
Thereafter, using this angle, the desired angular velocity $\omega_d$ of the underwater robot is calculated using~\eqref{eq:hsc}, which is provided to the velocity controller of the robot.

In this study, the gain $K_p, K_d$, which determines the magnitude of the haptic feedback, is set such that it can be easily reversed by the operator; hence, the robot rotates to the ideal posture even if the operator does not apply any force to the joystick in the lateral direction.

\section{EXPERIMENT}
\label{experiment}
\subsection{Scenario}

\begin{figure}[tb]
\centering
\includegraphics[width=1\linewidth]{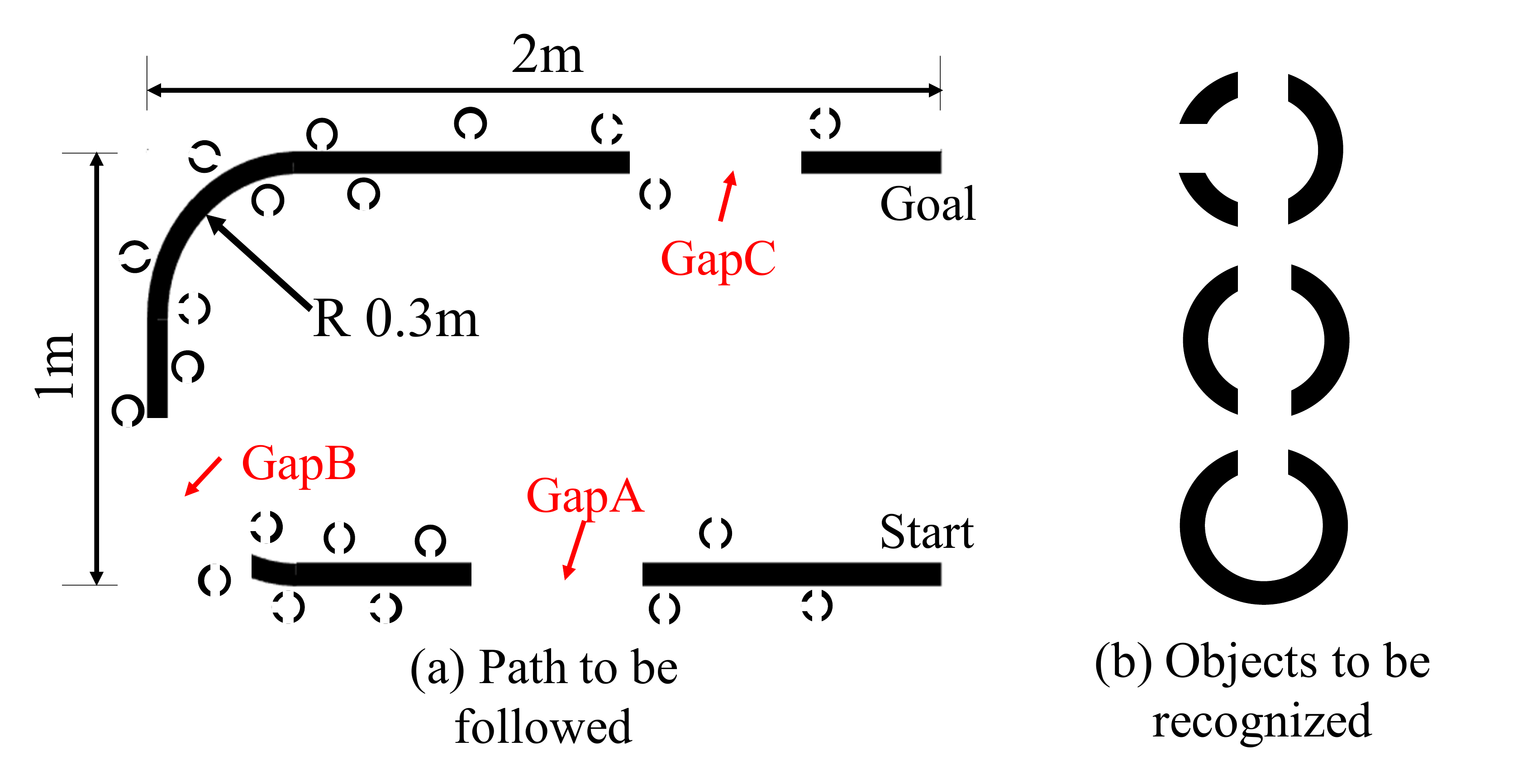}
\caption{Path and objects used in the experiments.}
 \label{fig:path and objects}
\end{figure}

In this study, we considered the case of performing an objects recognition task while following a path.

The objects depicted in Fig.~\ref{fig:path and objects} were placed on the water surface with a path to be followed, and several visual targets were placed around the path for the objects recognition task. The tracking error from the path was calculated using the images obtained from the camera attached to the top of the underwater robot. To reproduce the situation where it is difficult for AUVs to follow the path, the path was cut off at three points, which is depicted by Gaps A to C in Fig.~\ref{fig:path and objects}~(a).
For simplicity, we attached an AR marker at the start and end of each curve to determine the path curvature.
As shown in Fig.~\ref{fig:path and objects}~(a), the path starts from the bottom-right corner, goes straight to the right, and then goes straight again after the two curves.
Because the control in the horizontal plane was the main target in the present study, motion in the roll-pitch direction and depth were controlled to maintain the initial values by the inertial measurement unit and depth gauge built into the underwater robot.

\subsection{Design}
The control method was treated as a within-subject factor. Each participant performed a trial in the above-mentioned scenario using the manual control (MC) condition and the proposed cooperative path-following control (CC) condition in a crossover design, in which the order of the conditions was counterbalanced.

\subsection{Underwater robot}
\begin{figure}[tb]
\vspace{2mm}
\centering
\includegraphics[width=1\linewidth]{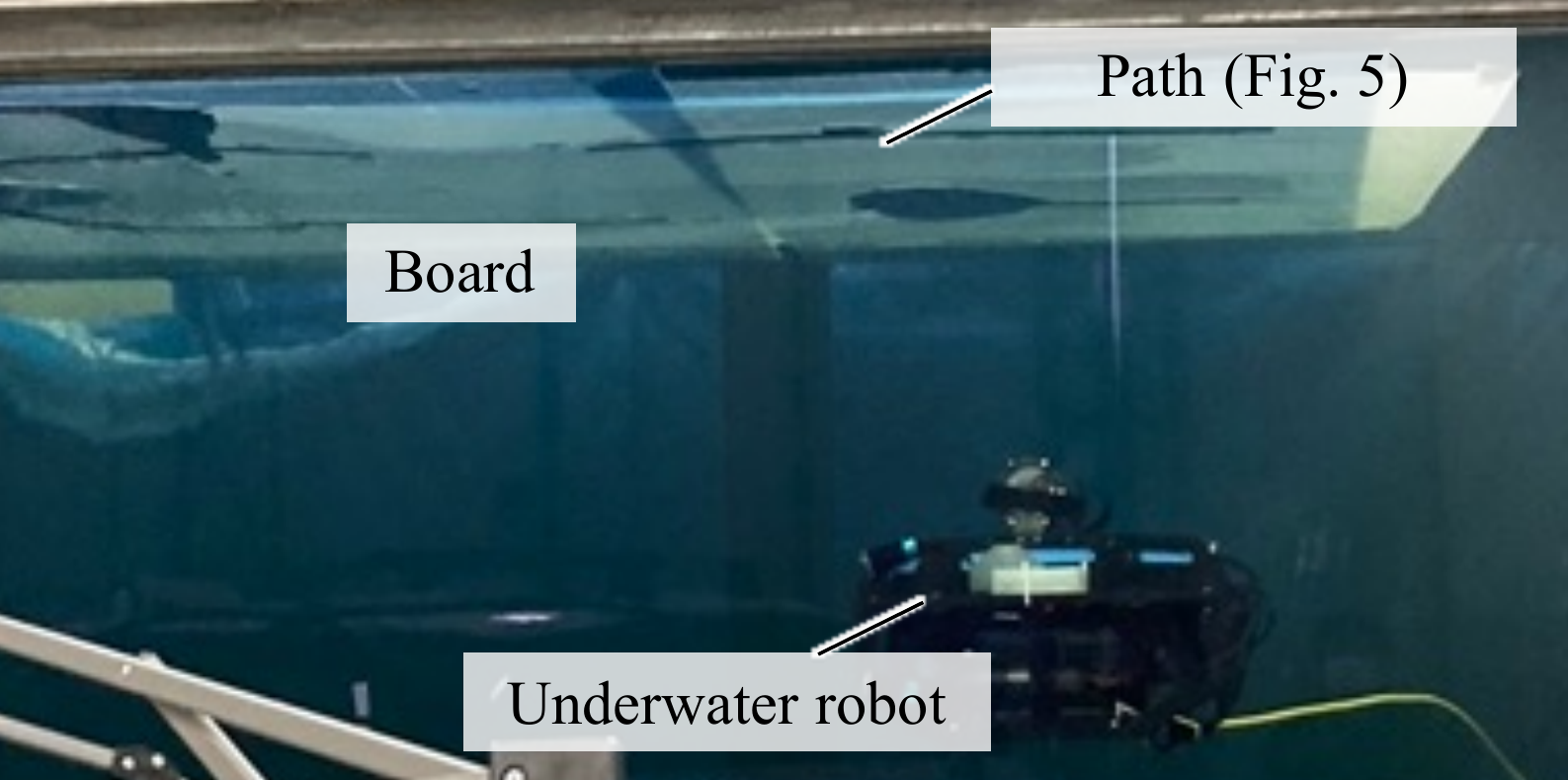}
\caption{Underwater robot in an indoor pool. }
 \label{fig:robot and pool}
\end{figure}
\begin{figure}[tb]
\centering
\includegraphics[width=0.97\linewidth]{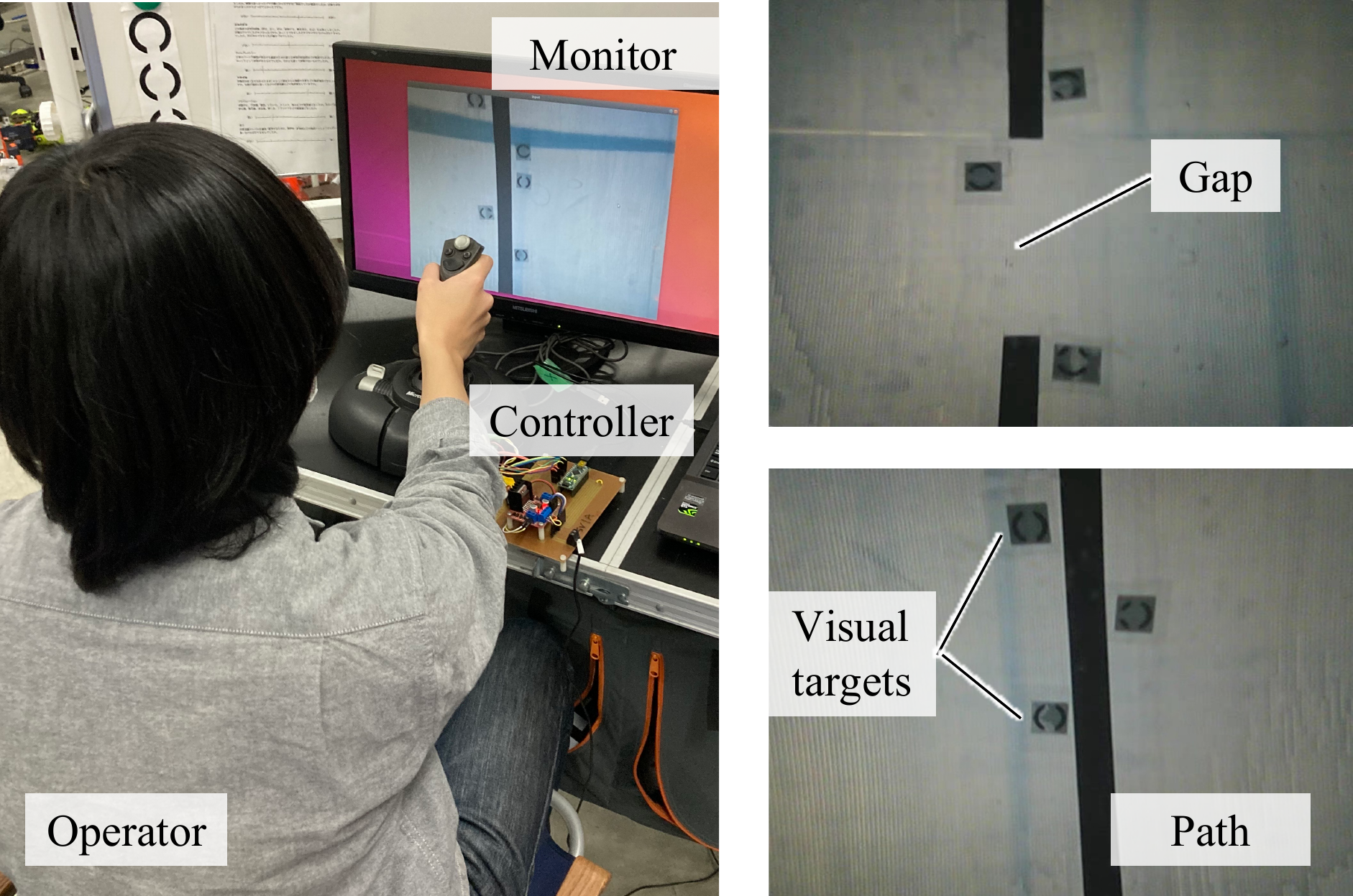}
\caption{Operator controlling the underwater robot and view of the operator.}
 \label{fig:operator}
\end{figure}

The proposed method was implemented on an underwater vehicle, as shown in Fig.~\ref{fig:robot and pool}, which was built by modifying the Blue ROV2 (Blue robotics) robot. A path-following experiment was conducted in an indoor pool~($5.4\times2.7\times1.6$ [m]) at Nara Institute of Science and Technology, Japan.
The underwater robot was equipped with four thrusters for the vertical motion and roll-pitch and four thrusters for motion in the horizontal plane and yaw.

\subsection{Participants}

Sixteen participants~(13 males and 3 females), who provided written informed consent, participated in the experiments.
All of them had no experience in operating underwater robots.

\subsection{Experimental Procedure}

Before the experiment, participants practiced the path-following task on straight and tortuous lines that were prepared only for the practice trial in the MC and CC conditions for 15 to 20 min each. 
In the practice trials, objects recognition tasks were not provided.
After the practice trials, the participants participated in the measurement trials using the image in the monitor (Fig.~\ref{fig:operator}), in which they were instructed to follow the U-shaped path, as shown in Fig.~\ref{fig:path and objects}~(a), while counting the number of circular visual targets with a pre-specified number of slits (Fig.~\ref{fig:path and objects}~(b)). The participants were also instructed to reach the goal as quickly as possible while giving the highest priority to object recognition.

After each measurement trial, the participants were asked to provide a subjective response to attention allocation and workload using the visual analog scale (VAS) developed for this research and the NASA-TLX~\cite{16_haga1996japanese}, respectively.

This experiment was conducted with the approval of the Research Ethics Committee of Nara Institute of Science and Technology (No. 2021-I-29).

\subsection{Evaluation Index}

The results of this experiment were evaluated using the following two evaluation methods:

1) Path-following performance (objective evaluation):\\
The path-following performance was evaluated using the root mean square error (RMSE) of the lateral deviation $e_2$ and angular deviation $e_3$ defined in Fig.~\ref{fig:model}. 

2) Attention allocation (subjective evaluation):\\
To evaluate the attention allocation of the operator, the participants were asked to answer the question “Which task did you pay attention to,” by a VAS, which consisted of two anchor words “path-following task” on the left and “objects recognition task” on the right.

3) Workload (subjective response):\\
The subjective workload during the task was evaluated using the Japanese version of the NASA-TLX~\cite{15_hart1988development,16_haga1996japanese}. 
Weighted workload (WWL) was evaluated as the weighted average of six subjective subscales: mental demand, physical demand, temporal demand, performance, frustration, and effort, ranging from 0 to 100; a higher score indicated a higher workload.

\section{RESULT}
\label{result}
In this experiment, the mean correct response rate for objects recognition by participants was 94~[\%] and 96~[\%] under the MC and CC conditions, respectively, and no statistical significance was found ($p=0.37$).
The mean times required to follow the path were 41~[s] and 44~[s] for the MC and CC conditions, respectively, and no statistical significance was found ($p=0.35$).

\subsection{Path-following task}
\begin{figure}[tb]
\vspace{2mm}
\centering
\includegraphics[width=1\linewidth]{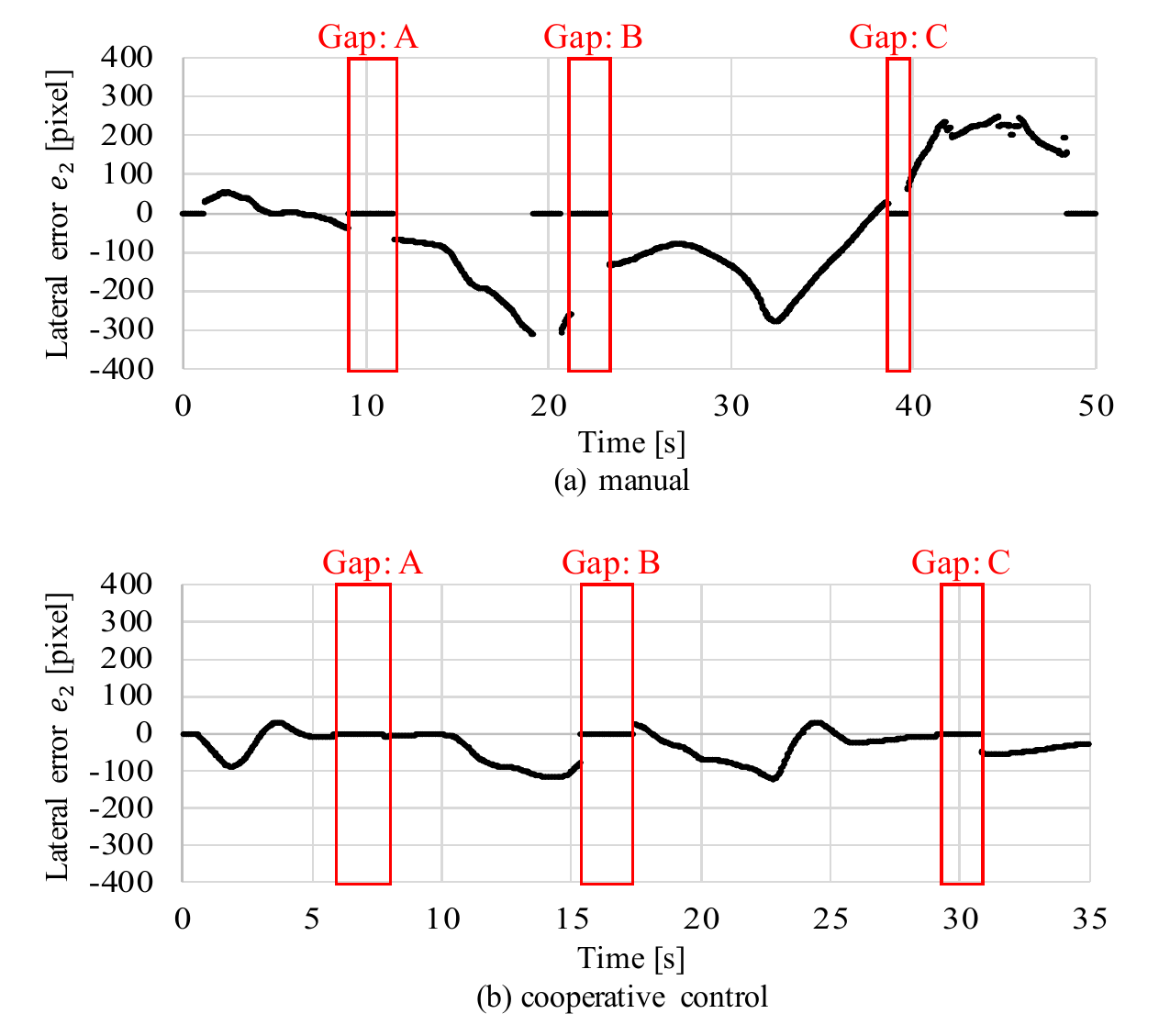}
\caption{Tracking error: lateral error.}
 \label{fig:9}
\centering
\includegraphics[width=1\linewidth]{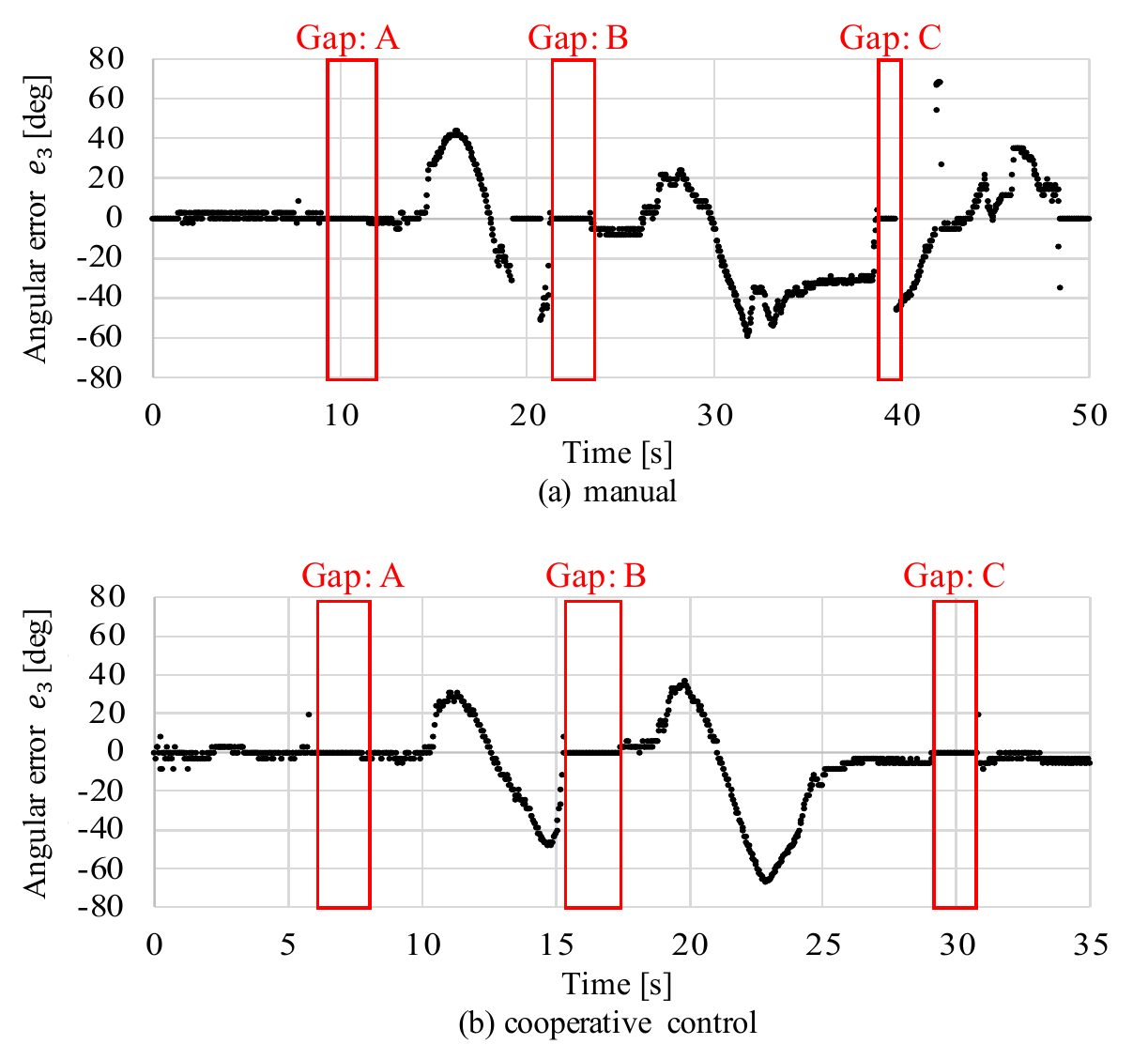}
\caption{Tracking error: angular error.}
 \label{fig:10}
\vspace{-3mm}
\end{figure}

Fig.~\ref{fig:9} and Fig.~\ref{fig:10} show an example of the time series data of the lateral error $e_2$ and angular error $e_3$, respectively, as defined in Fig.~\ref{fig:model}.
In both figures, the upper and lower graphs show under the MC condition and the CC condition, respectively. 

The larger lateral error was found under the MC condition than under the CC condition. 
In particular, the path was lost from the camera image at approximately 20~[s] under the MC condition. 
However, under the CC condition, the error that occurred when the underwater robot reached the curves was immediately corrected by the operator.

In addition, a smaller angular error was found under the CC than under the MC, and the operator could immediately correct the angular error after the curves under the CC condition.
Fig.~\ref{fig:12}~(a) and Fig.~\ref{fig:12}~(b) show the RMSE of the lateral and angular deviations.
For the lateral error, the Shapiro-Wilk test~(python2.7) and F-test revealed that the null hypotheses of normality (MC: $p=0.11$, CC: $p=0.34$) and equal variance~($p=0.29$), respectively, were not rejected. The paired t-test revealed that the lateral error under the MC condition was significantly larger than that under the CC condition ($p=0.000$).

For the angular error, the Shapiro-Wilk test and F-test showed that the null hypotheses of normality (MC: $p=0.35$, CC: $p=0.89$) and equal variance ($p=0.090$), respectively, were not rejected. The paired t-test revealed that the angular error in MC was significantly larger than that in CC ($p=0.039$).

\subsection{Attention allocation to objects recognition}

Fig.~\ref{fig:14}~(a) shows the results of the subjective responses of the attention allocation of the participants to the objects recognition task or the path-following task.
A higher value indicates that more attention was allocated to the objects recognition task than the path-following task of the robot.
The Shapiro-Wilk test showed that the null hypothesis of normality was rejected (MC: $p=0.040$, MC: $p=0.0034$). 
The Wilcoxon signed rank-test showed that the attention allocation of the participants was not significantly changed among the method conditions ($p=0.13$), although the median under the CC condition tended to object recognition.

\subsection{Total Workload}

\begin{figure}[tb]
\vspace{2mm}
\centering

\centering
\includegraphics[width=1\linewidth]{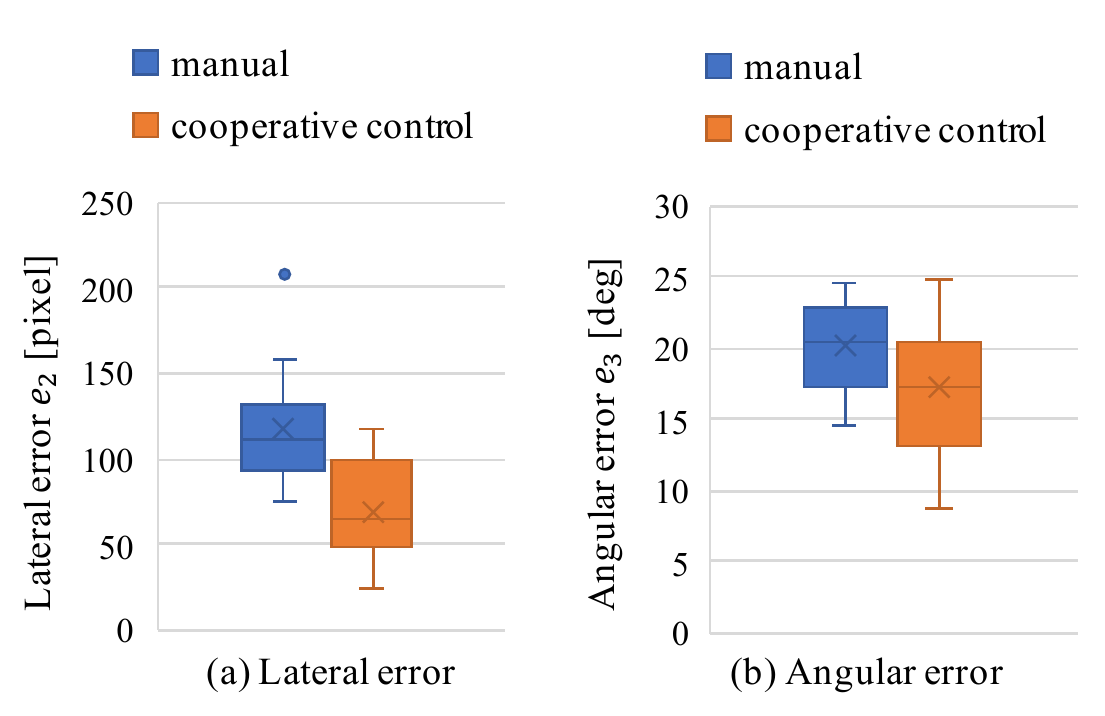}
\caption{Tracking errors (RMSE).}
\label{fig:12}
\centering
\includegraphics[width=1\linewidth]{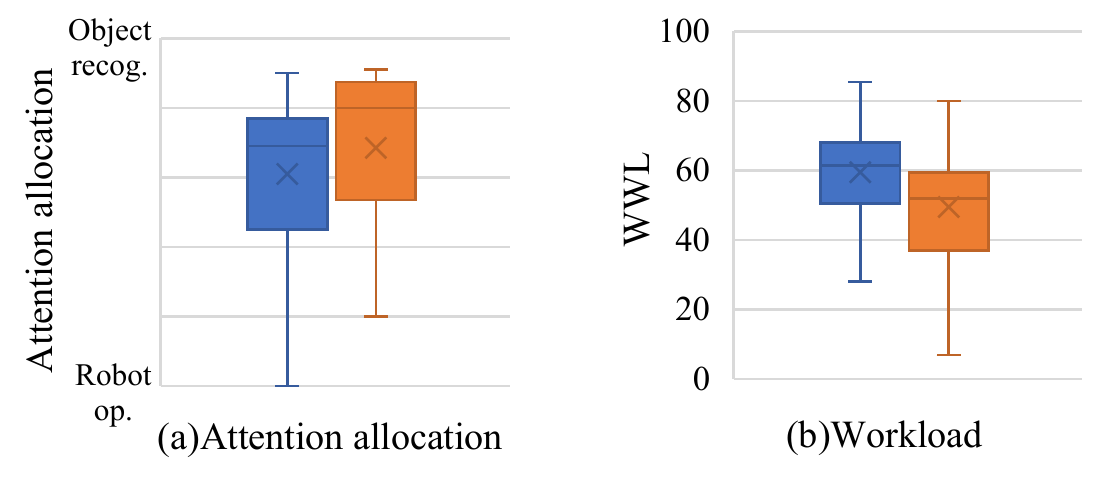}
\caption{Subjective evaluation.} 
\label{fig:14}
\vspace{-4mm}
\end{figure}

Fig.~\ref{fig:14}~(b) shows the WWL scores of the NASA-TLX.
The null hypotheses that normality (MC: $p=0.74$, MC: $p=0.53$) and equal variance ($p=0.29$) between method conditions were not rejected.
The paired t-test showed that WWL in the CC condition was significantly smaller than that in the MC condition ($p =0.044$).

\section{DISCUSSION}
\label{discussion}
A cooperative path-following controller for objects recognition tasks while following a path using an underwater robot was proposed. 
The results of the experiments showed that the proposed method significantly reduced the tracking error in the lateral position, the posture angle, and the subjective workload compared to the conventional method, even in a situation where the path was interrupted in the middle of the path and should be handled with only manual control without any assistance from the automatic control. The correct recognition rate of the objects recognition task and task execution time did not change significantly under the control method condition. 
This result suggests that the support of the proposed method facilitated the path-following by the operator, resulting in comparable task performance with less workload than with manual control.

The improvement in the path-following performance under CC condition was more apparent in the lateral direction. 
This is because of the proposed concept of the human-machine collaboration method, in which the operator entrusts lateral control as much as possible when the robot functions correctly and humans are allowed to intervene in the posture control of the robot, which is more important for the objects recognition task. 
Given the proposed method, the operator can directly intervene in the robot posture by the lateral joystick input and indirectly intervene in the longitudinal direction, which is controlled by the translational velocity conversion through posture control by the lateral direction operation of the joystick. 
It should be noted that the application of the CC tends to make the operator devote more attention to object recognition (Fig.~\ref{fig:14}~(a)), although the difference is not statistically significant. 
In the part where the path was not broken, the operator only needed to operate the joystick in the longitudinal direction, and the underwater robot could follow the path with the assistance of the system, while in the part where the path was broken, the operator needed to pay attention to the path-following task as well.
The workload was significantly reduced by the proposed system, however, the workload increased under the CC condition for three of the sixteen participants.
One possible explanation could be that the strength and timing of the system assistance were different from what the operator expected, although no clear explanation could be given. Understanding and resolving such human-machine conflicts\cite{saito2018control, okada2020transferring} is an important issue that needs to be addressed in the future.

Several studies have shown that HSC can improve the performance of vehicle motion control through human-machine cooperation and reduce the workload. The results of the present study illustrate the same trend. 
For example, the introduction of HSC to the lane-keeping control of a surface vehicle has been shown to improve lane-following performance and overall performance by reducing the workload \cite{06_mulder2012sharing,05_nishimura2015haptic}. 
In addition, in controlling the movement of deep-sea vehicles to avoid obstacles, the path-following performance between obstacles has been improved \cite{wang2014haptic}.
We studied the cooperative control of human and underwater robots using HSC. For example, Suka~\cite{Suka2022} et al. proposed a cooperative control method based on a simple proportional-derivative (PD) control law that performs objects recognition tasks while maintaining a stationary position and confirmed that it is effective in reducing the operator workload.
In addition, in~\cite{08_konishi2020haptic}, we conducted a study on a task similar to that presented in the current study. 
In the previous study,~\cite{08_konishi2020haptic}, a nonholonomic underwater robot that does not move directly in the lateral direction, was considered and the longitudinal and rotational motions of the robot were assigned to two-dimensional joystick motion, and HSC was introduced in the lateral direction to reduce the operator workload in the path-following task. 
However, a simple control method using preview control, in which PD feedback for the predicted error in the future path was used, was applied; but the convergence of the tracking error could not be guaranteed theoretically.
Based on the results of this preliminary study \cite{08_konishi2020haptic}, the present study proposes a new cooperative path-following control method suitable for human-machine cooperation by designing an inverse optimal control for a holonomic underwater mobile robot. 
Using the proposed method, we demonstrated an improvement in the path-following performance and a reduction in the workload of the operator. 
The fact that the improvement in path-following performance was confirmed in the present study, which was not observed in the previous study \cite{08_konishi2020haptic}, can be regarded as the effectiveness of the proposed control law, although it is difficult to directly compare the results of the preliminary study and the present study owing to differences in mechanisms, including degree of freedom.
It should also be noted that the control law proposed in this study can be easily applied to stationary control, as reported in \cite{Suka2022} to achieve better control.

When something is mistakenly recognized as a path, the proposed method may provide incorrect guidance. 
The $\beta$ used for the translational velocity conversion cannot be directly changed by the operator, but the arbitrary motion of the robot can be achieved through posture change by the lateral joystick input. 
If the operator notices a misrecognition of the robot, the guidance can be temporarily stopped by the pressing of a button. 
In the present study, we demonstrated that the system can switch off the guidance control by setting $e_2$ and $e_3$ to zero when the system judges that the path detection has failed so that the robot can transfer the control to manual control and continue the task without any problem. 
It should be noted that a control discontinuity may occur when the assistance once turned off is turned back on, which might lead to an unstable phenomenon. A smooth control transition method using HSC has been proposed to eliminate control discontinuities in the control transition between the automated driving of automobiles and manual driving \cite{saito2018control,okada2020transferring}, which is expected to be applied to the problem.

\section{CONCLUSION}
\label{conclusion}
\subsection{Contribution}
In this study, we proposed a new collaborative path-following control method of underwater robots for visual objects recognition tasks while following a path to reduce the operator workload. 
In the proposed method, the design concept of the human–machine interface, including translational velocity conversion and haptic guidance for rotational motion, is combined with a path-following controller based on inverse optimal control.
The subject experiment verified the effectiveness of the proposed method from viewpoints of task performance and operator workload; the path-following performance was significantly improved and the subjective workload was significantly reduced under the effect of the proposed method in comparison to manual control, even though the robot occasionally failed to detect the followed path.

\subsection{Future research direction}

In this study, a path-following problem was limited to a two-dimensional plane. 
Expansion of the proposed method for six-DoF motions in a three-dimensional space is an important future study.

In addition, the effectiveness of the proposed control law was verified in an indoor pool without a water flow. 
Thus, conducting experiments in the presence of external disturbances, such as flow, especially in real environments, such as oceans and lakes, is also important in the future course of action.

\section*{ACKNOWLEDGEMENT}
This research was partially supported by JSPS KAKENHI Grant Number JP21H01294, Japan.

\bibliographystyle{ieeetr}
\bibliography{sample.bib}

\end{document}